\documentclass{article}

\PassOptionsToPackage{numbers, compress}{natbib}

\usepackage[final]{neurips_2024}

\usepackage[utf8]{inputenc} 
\usepackage[T1]{fontenc}    
\usepackage{hyperref}       
\usepackage{url}            
\usepackage{booktabs}       
\usepackage{amsfonts}       
\usepackage{amsmath}
\usepackage{nicefrac}       
\usepackage{microtype}      
\usepackage{xcolor}         
\usepackage[pdftex]{graphicx}

\title{Stylish and Functional: Guided Interpolation Subject to Physical Constraints}

\author{%
  Yan-Ying Chen$^*$, Nikos Arechiga\thanks{Equal contribution. Work was done at Toyota Research Institute.} , Chenyang Yuan, \\ \textbf{Matthew Hong, Matt Klenk, Charlene Wu}\\
  Toyota Research Institute\\
  Los Altos, CA\\
  \texttt{yan-ying.chen@tri.global}
}

\begin{document}

\maketitle

\begin{abstract}
Generative AI is revolutionizing engineering design practices by enabling rapid prototyping and manipulation of designs. One example of design manipulation involves taking two reference design images and using them as prompts to generate a design image that combines aspects of both.
Real engineering designs have physical constraints and functional requirements in addition to aesthetic design considerations.
Internet-scale foundation models commonly used for image generation, however,
are unable to take these physical constraints and functional requirements into consideration as part of the generation process.
We consider the problem of generating a design inspired by two input designs, and propose a zero-shot framework toward enforcing physical, functional requirements over the generation process by leveraging a pretrained diffusion model as the backbone. As a case study, we consider the example of rotational symmetry in generation of wheel designs. Automotive wheels are required to be rotationally symmetric for physical stability. 
We formulate the requirement of rotational symmetry by the use of a symmetrizer, and we use this symmetrizer to guide the diffusion process towards symmetric wheel generations.
Our experimental results find that the proposed approach makes generated interpolations with higher realism than methods in related work, as evaluated by Fréchet inception distance (FID). We also find that our approach generates designs that more closely satisfy physical and functional requirements than generating without the symmetry guidance.
\end{abstract}

\section{Introduction}
\begin{figure}[!htbp]
    \centering
    \includegraphics[width=1\textwidth]{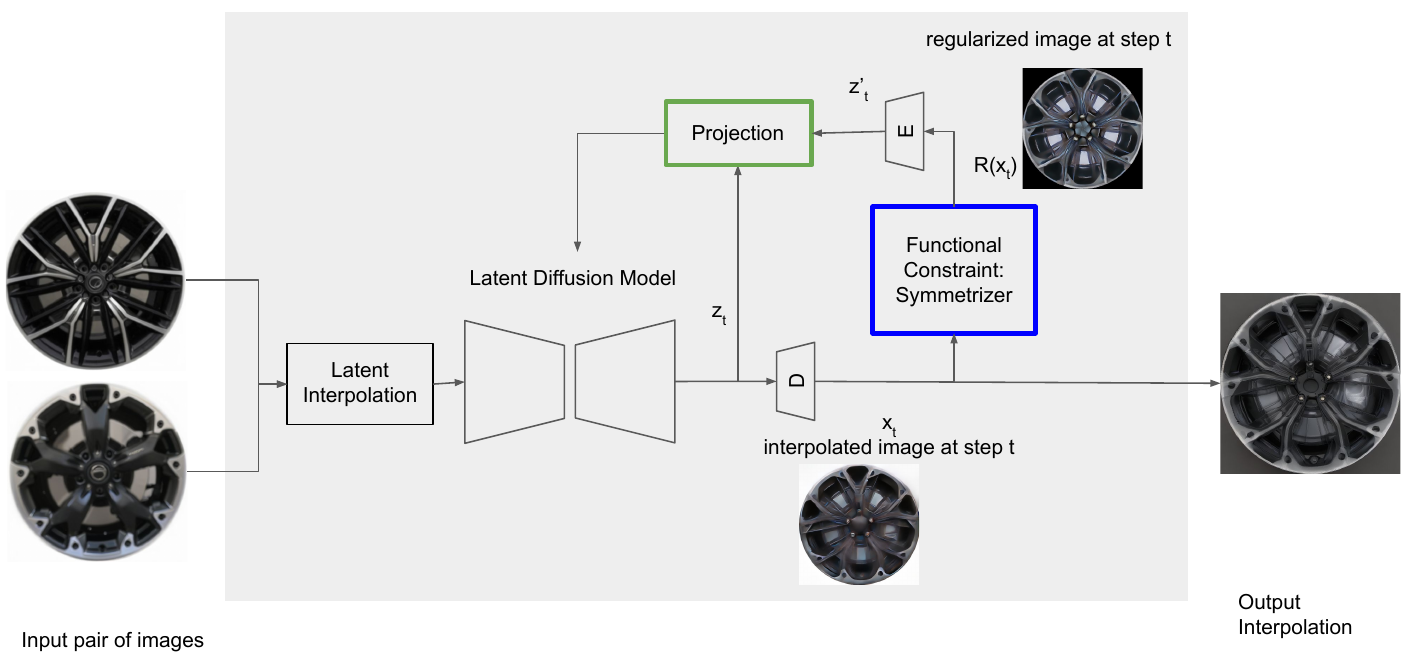}
    \caption{System overview: The proposed model takes a pair of reference images as the input and generates an intermediate interpolation, where the interpolation is gradually regularized with functional constraints in each step of the denoising process.}
    \label{fig:system}
\end{figure}
Generative models (\cite{rombach2022high, pmlr-v139-ramesh21a, karras2019}) are reshaping design processes by significantly reducing the time and effort required to create visual content. This trend pushes the effort of grounding the generative models in the perspectives of design thinking \cite{designcouncil2005} (e.g., generating diverse designs to support exploration \cite{cai2023}, converging ideas with potential variations such as combinations between different references \cite{davis2023}).

Applying generative AI to real-world product engineering design, such as a car wheel, requires a careful balance between creativity and practicality. While generative models can traverse an embedding space of image patterns to output unique visual styles, their application must be compatible with existing design workflows and adhere to strict physical and functional requirements \cite{hagtvedt2014}.

Interpolated combinations of reference designs is a technique well-aligned with a prominent design intent and practice, that can also provide creative surprises \cite{jeon2024}. Prior work has considered blending images using traditional image processing approaches \cite{wolberg1998image,li2017pixel} and more recently has leveraged deep representation learning and generative models \cite{karras2019,Zhang_2020_WACV,wang2023interpolating}. Karras et al. \cite{karras2019} takes two different samples to the synthesis network at different levels to mix styles of the two. Wang et al. \cite{wang2023interpolating} interpolates reference images in latent space at a sequence of decreasing noise levels to generate interpolations with diffusion models without additional training data. Research in generative interpolation also addresses smoothness of transition between sequential interpolations such as DiffMorpher \cite{zhang2024diffmorpher} and Smooth-Diffusion \cite{guo2024smooth}.

\begin{figure}
    \centering
    \includegraphics[width=0.5\textwidth]{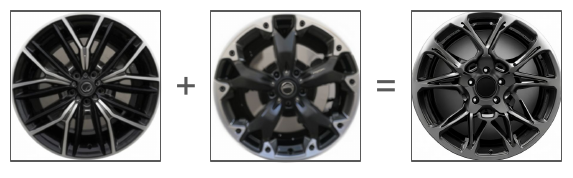}
    \caption{Generating a interpolated combination of two different image sources can create distortion that poses critical challenges to engineering functionality. For example, the distorted spokes are not rotationally symmetrical and hence may distribute force unevenly and affect smooth rolling.}
    \label{fig:distortion}
\end{figure}

However, there has not been much work on blending images that represent engineering designs, subject to realism as well as functional constraints. This is critical because blending different sources of designs can easily create distortion and pose challenges to the generated design to perform the intended functionalities of the product. Figure \ref{fig:distortion} shows examples of interpolated wheels with distortion that makes it unlikely to function as wheels.

We propose a framework of enforcing functional constraints in interpolation generation by fusing generated interpolation and its regularized form in the denoising steps of latent diffusion model. The regularizer constrains a certain functionality of the generated interpolations such as symmetry. We propose two symmetrizers as example constraint regularizers for guiding car wheel generation. The experiment results show that the proposed approach improves the realism and diversity (lower FID score) of the generated interpolations and make the interpolations better conform to the functional constraint of symmetry, as measured by the quantitative symmetry metrics. The framework does not require additional training, which significantly reduces the labeling efforts. 

Our contributions includes: (1) a framework towards enforcing physical constraints and functional requirements in the denoising process of interpolation generation; (2) two example regularizers of functional constraints in terms of symmetry; (3) a weighting mechanism to manage the tradeoff between realism and regularization in the projection.


\section{Related Work}
Deep generative models such as generative adversarial networks and diffusion models can catch the underlying distribution of the training dataset and synthesize plausible and realistic images. Variations of these models have been used to generate interpolated combinations of given reference images by mixing their latent representations. The style mixing in StyleGAN \cite{karras2019} employs mixing regularization to allow two latent codes through the mapping network to generate an image. However, GAN based approaches may suffer from challenges such as mode collapse \cite{roth2017,jabbar2021} and require a inversion process \cite{xia2022} to map a given real image to the latent code in the latent space, which allows an image to be faithfully generated from the inverted code. 

Latent diffusion models (LDM) \cite{rombach2022high} are one of the state-of-the-art models for image generation that has a more stable training process without the concern of mode collapse and is known for generating high quality images. LDM has been used for generating combined interpolations \cite{guo2024smooth,zhang2024diffmorpher} for image morphing with smooth transition. Wang et al. \cite{wang2023interpolating} has leveraged the LDM architecture to create a model for interpolating images in the latent space at a sequence of decreasing noise levels, and explore the way to manipulate the generated interpolations using text and human pose as constraints. 

The advantages of using diffusion models as a backbone are (1) accommodating control guidance such as text and image as constraints and (2) building upon strong pretrained models that allows finetuning an transfer learning \cite{Ruiz_2023_CVPR, hu2022lora,xucvpr2024} as well as enabling zero-shot methods \cite{pmlr-v139-ramesh21a,choi2021}. Prior work has extended the control guidance to a variety types of constraints. ControlNet \cite{zhang2023} reuses a pre-trained diffusion model to learn diverse controls such as edge, pose and segmentation maps. However, it requires an adequate amount of training images paired with corresponding controls. Null space approach \cite{pmlr-v139-ramesh21a} adds linear constraints to diffusion models in a zero-shot setting.
However, these controls have not covered the functional constraints required in engineering design and have not been applied to interpolation generation. 

Our proposed interpolation model uses LDM as the backbone and integrate the functional constraints in the noise schedule at the inference stage. This allows guiding the interpolation subject to functional constraint in a zero-shot setting. This is critical for engineering design use cases where data scarcity could pose challenges for training and functional constraints may be specific to a particular surrogate model that are not supported by general vision tasks such as edge detection and pose estimation. 

\section{Problem setting}
Given a pair of two physical object images, we want to generate a third image that blends the style and semantic properties of the two inputs, while respecting a constraint on the generated object. As a case study, we consider the problem of taking two wheel images and blending them to produce a wheel that respects the constraint of rotational symmetry, which is an essential property of functional wheels.

\section{Proposed method}
The proposed system encodes \textbf{F}unctional Constraints in \textbf{I}n\textbf{T}erpolation (\textbf{FIT}). FIT includes three major components: (1) latent interpolation, (2) functional constraint regularization, using a symmetrizer as an example case and (3) projection to integrate both the generated image and the regularized image at each time step of the denoising process.

\subsection{Latent interpolation} \label{sec:latent_interpolation}
The backbone of the generated interpolation is based on Latent Diffusion Models \cite{rombach2022high}. Given a pair of input images, we first encode each image to a latent vector. Following the latent interpolation strategy suggested in \cite{wang2023interpolating}, both latent vectors are added with shared noise and then interpolated using spherical linear interpolation. We then denoise the interpolated latent vector $z$ using DDIM sampling \cite{song2021denoising} to generate an intermediate image. We adopt the noise schedule used in \cite{wang2023interpolating} in our implementation. Different from \cite{wang2023interpolating} that generates interpolation at the end of DDIM sampling, we propose to regularize the latent interpolation in each step of the DDIM sampling that enforces the designated functional constraint to the generated interpolations. 
We describe the functional constraint of our case study in Section \ref{sec:functional_constraint}.

\subsection{Functional Constraint Regularization} \label{sec:functional_constraint}
We consider two formalizations of our functional specification: (1) rotate and average, and (2) select sector and stitch.

\noindent {\bf Rotate and Average (RA):} Given a wheel with a pattern that repeats $n$ times, let $\{A_k\}_{k=1}^n$ be the set of $n$ rotation matrices, such that each $A_k$ represents a rotation by $2 \pi /k$. Then, an image $x$ has $n$-fold rotational symmetry if for all $i,j \in [n]$,  $A_i x = A_j x$. We can project any given image $x$ to this linear subspace of rotationally symmetric images by computing $x^* = \frac{1}{n} \sum_{k=1}^n A_k x$.

\noindent {\bf Select sector and stitch (SS):} Let $\{B_n\}$ be the set of matrices that select one of the $n$ radial sectors of the image, i.e. a matrix populated with ones for the pixels corresponding to the relevant sector and zeros everywhere else. An image $x$ has n-fold rotational symmetry in this formalization if all of its sectors are equal, i.e. if $B_i x = B_j x$ for all matrices $B_i$, $B_j$. Given an image $x$, we can compute its symmetric projection by choosing a sector and assigning all of the other sectors equal to it.

\subsection{Projection} \label{sec:pooling}
We propose to integrate an intermediate interpolation with its regularized form at every step of the denoising process, where the regularized form is computed according to the functional constraint regularization to obtain a projection to the set of rotationally symmetric images. The integration considers two factors: (1) the difference incurred after regularization and (2) the decay of impact from the regularization over generation steps.

As presented in Figure \ref{fig:system}, in each step $t$ of DDIM sampling the latent interpolation (Sec. \ref{sec:latent_interpolation}) $z_t$ is decoded into $x_t$, which is then regularized by a functional constraint $R(\cdot)$ such as RA or SS introduced in Sec. \ref{sec:functional_constraint}. The regularized image $R(x_t)$ is encoded as $z^r_t$, then pooled with $z_t$ to obtain $z'_t$:
\begin{align}
x_t &= D(z_t) \\
z^r_t &= E(R(x_t)) \\
\lambda &= s(z_t,z^r_t) \frac{w}{t^d}, \label{eq:weighting} \\
z'_t &= (1-\lambda){\cdot}z_t+\lambda{\cdot}z^r_t
\end{align}
where $s(z_t,z^r_t)$ is the cosine similarity of $z$ and $z^r_t$, and is normalized to $0$ to $1$. $w$ is a weight set to adjust the impact from $z^r_t$ and $d$ is a constant set to control the decay speed of that impact. $\lambda$ is subject to the similarity of $z_t$ and $z^r_t$ to ensure the the pooling would not induce too much noise incurred by the difference between the original image and the regularized image. The impact from the regularized image is decayed over sampling steps to decrease the intervention to the denoising process which could create additional noise and decrease realism, especially toward the end of the generation process. The ablation study for similarity, decay and weight is reported in Sec. \ref{exp:discussion}.

Our work bears similarities to the work on null-space diffusion of \cite{wang2023nullspace}, but differs in the introduction of a decayed pooling between the regularized image and the original, unregularized image -- in the context of generating an interpolated image given a pair of reference images.


\section{Experiments}
We conducted experiments to answer the following research questions: (1) Does FIT generate interpolations close to real designs? (\ref{exp:realism}) (2) Does FIT generate interpolations that conform to the specified functional constraint? (\ref{exp:symmetry}) For the first question, we measure the Fréchet inception distance (FID) \cite{heusel2017} between generated wheel interpolations and real wheels. For the second question, we measure the symmetry score as subject to the symmetrizing functions used in generating interpolations. 

 We measure a symmetry score by comparing the differences between a generated interpolation and its symmetrized form. Since we consider two different ways to compute the symmetrized image, we report two different symmetry scores. Using the notation of Section \ref{sec:pooling}, $R(x_t)$ is the regularization of image $x_t$, which may be computed by either SS or RA. Then, our symmetry score is computed as
$$
\text{Sym} = 1 - \frac{\left|\left|R(x_t) - x_t\right|\right|}{K}
$$
where $K$ is a normalization constant equal to the square root of the product of dimensions of $x_t$. 

\subsection{Experiment Setting}
To align with real design use cases, we use a commercial grade vehicle image dataset from EVOX \footnote{The images used in this study are the property of EVOX Productions, LLC and are subject to copyright law. The appropriate licenses and permissions have been obtained to ensure the rightful use of these images in our study. For more information, visit \url{https://www.evoxstock.com/}.} as the experiment data. The dataset includes high-resolution images across multiple car models collected in year 2022 and 2023. We use a wheel detector trained by YOLO8 \footnote{\url{https://github.com/ultralytics/ultralytics}} to extract wheels from the car images. Each wheel is located in the center of an image, and the brand logo in the center is blurred. Each image is normalized to $512\times512$ pixels. We first sample 380 different pairs of original wheel images from this dataset as the parent images. Similar to the setting in \cite{wang2023interpolating}, each pair of parent images generates one interpolated image, where the generated image and a parent image are then used to generate another interpolated image in a branching structure. In total, 5,700 interpolated images are generated for evaluation. $50\%$ of interpolations are used as the validation set and the rest are used as the test data. Note that, the validation set is used to determine the hyperparameters such as symmetrizing weight ($w$), the use of decay ($d$) and similarity ($s(z_t,z^r_t)$), and the same parameters are applied to the test set. The image prompts for generating the interpolations in the validation set and the test set are separate. The baseline is the diffusion based interpolation model \cite{wang2023interpolating} without considering functional constraints. Both FIT and the baseline use Stable Diffusion 1.5 as the backbone models with the same parameter setting.


\subsection{Realism of Generated Interpolations} \label{exp:realism}
Distortion can occur in generated interpolations and hence affect functionality. We evaluate whether enforcing functional constraints can reduce the distortion and increase realism. We use FID to evaluate the generated wheel interpolations against real wheels in our dataset. The entire set of 1,439 real wheel images are used to be compared with interpolations for measuring FID. As reported in Table \ref{tbl:fid}, FIT obtains lower FID than the baseline using either functional regularizer SS or RA. This indicates that enforcing functional constraint would help generate more realistic engineering designs.

\begin{figure}
    \centering
    \includegraphics[width=0.7\textwidth]{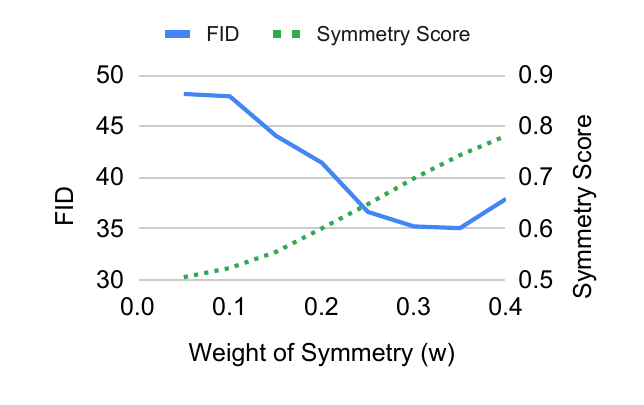}
    \caption{FID and symmetry score of the generated images over different weight of symmetry (w) in Eq. \ref{eq:weighting} that adjusts the impact from the regularization.} 
    \label{fig:weight}
\end{figure}

In addition, the strength from the functional constraints (controlled by the weight $w$ of symmetrizing) can affect the realism of the generation. In Figure \ref{fig:weight}, as we increase the weight from 0.1, the FID improves to a point, but overweighting results in less realism. That is, increasing the weight creates additional noise that can be a burden for generating realistic designs, and, in our approach, there exists a trade-off between strong functional compliance (symmetry) and realism.

\begin{table}
  \caption{FID of images generated by the baseline and the proposed approach in the validation and test sets. (lower is better)}
  \label{tbl:fid}
  \centering
  \begin{tabular}{lll}
    \toprule
    Approach     & FID (Val.)$\downarrow$     & FID (Test)$\downarrow$  \\
    \midrule
    (Wang 2023) & 39.28  & 46.01     \\
    FIT (SS)    & 33.98 & 41.44      \\
    FIT (RA)    & \textbf{30.85} & \textbf{33.38}  \\
    \bottomrule
  \end{tabular}
\end{table}

\subsection{Symmetry of Generated Interpolations} \label{exp:symmetry}
Another goal of FIT is to increase functional compliance of the generated interpolations. Taking the functional constraint SS in Sec. \ref{sec:functional_constraint} as an example, the generated interpolations should gradually conform to that desired constraint. 

\begin{figure}
    \centering
    \includegraphics[width=1\textwidth]{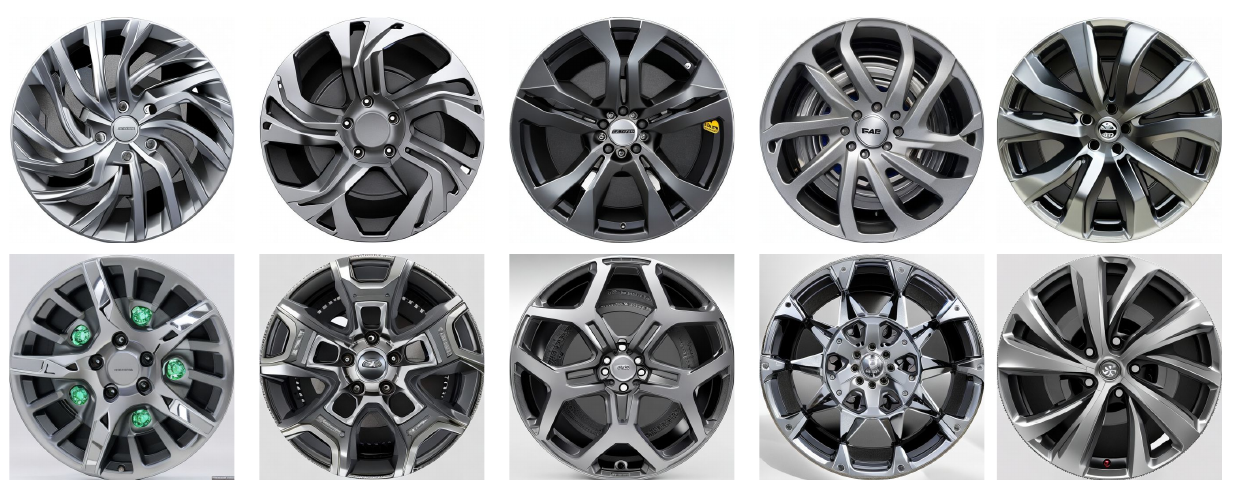}
    \caption{Examples of images generated by the baseline \cite{wang2023interpolating} (top row) and FIT (bottom row).}
    \label{fig:generated_wheels}
\end{figure}

The smaller the differences between an image and its regularized version, the higher the symmetry score. As reported in Table \ref{tbl:sym_score}, FIT can generate wheel interpolations with higher symmetry scores than the baseline, indicating that the interpolations better conform to a desired functional constraint such as symmetry. Example interpolations generated by FIT and the baseline are presented in Figure \ref{fig:generated_wheels} for a qualitative comparison of symmetry.


\begin{table}
  \caption{Symmetry scores of the images generated by the baseline and the proposed approach. Higher is better Sym (SS) and Sym (RA) are the symmetry measures based on the regularizer SS and RA, respectively.}
  \label{tbl:sym_score}
  \centering
  \begin{tabular}{lll}
    \toprule
    Approach     & Sym (SS)$\uparrow$     & Sym (RA)$\uparrow$  \\
    \midrule
    (Wang 2023) & 0.507  & 0.882     \\
    FIT (SS)    & \textbf{0.601} & 0.867     \\
    FIT (RA) & 0.519 & \textbf{0.909} \\
    \bottomrule
  \end{tabular}
\end{table}

\subsection{Discussion} \label{exp:discussion}
\textbf{Similarity and Decay} -- While regularization increases the compliance, we notice that the additional noise from regularization can reduce the realism, especially if the noise is large or is added toward the end of generation. FIT proposes two strategies to address this issue. First, the strength from the regularization is subject to the similarity of the generated image in each DDIM step and its regularized form. The lower similarity, the higher induced noise and hence a smaller strength from the regularization is given to the pooling process. Table \ref{tbl:ablation_study} shows that considering similarity (similarity \& decay) obtains better FID (more realistic) than without considering similarity. The second strategy is to reduce the strength of regularization over DDIM steps. Without decaying the strength, the FID can be dramatically increased as reported in Table \ref{tbl:ablation_study} while using the decay mechanism reaches a much lower FID. 

\begin{figure}[!htbp]
    \centering
    \includegraphics[width=0.4\textwidth]{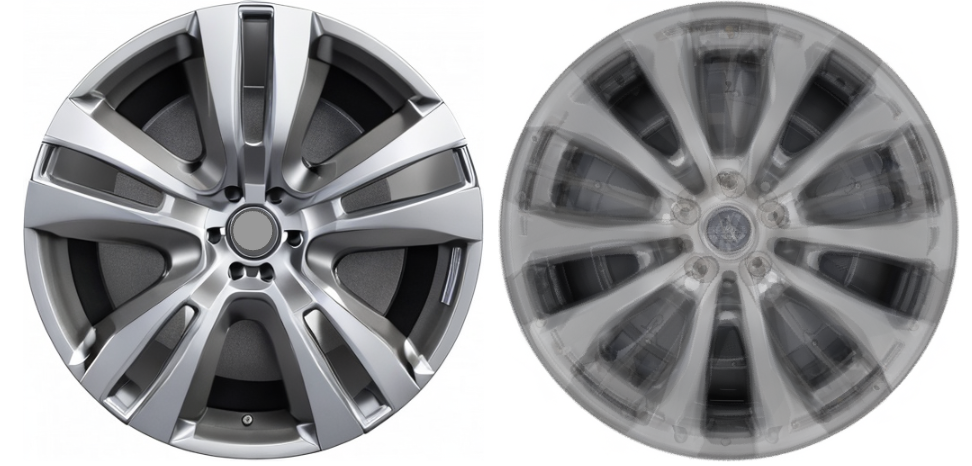}
    \caption{Example images generated by FIT (RA): Left: Applying the constraint during interpolation generation. Right: Applying the constraint at the end of interpolation.
}
    \label{fig:oneshot}
\end{figure}

\textbf{Constraint in the middle or at the end} -- We also investigate whether interleaving generation and regularization over steps is better than imposing constraint to a generated image at the end \cite{wang2024idetc}. The results in Table \ref{tbl:run_last} shows that it may depend on the characteristics of the desired functional regularizer. Some regularizer such as RA can induce a considerable amount of noise that reduces realism (Fig. \ref{fig:oneshot} right), and thus interleaving generation and regularization allows further denoising to increase the realism (lower FID).  

\begin{table}
  \caption{Ablation study for considering (1) the similarity of the intermediate interpolation and its regularization form and (2) the decay of impact from regularizer over denoising steps (Sec. \ref{sec:pooling}. Considering both consistently performs the best when using different initial weight of symmetry ($w$).}
  \label{tbl:ablation_study}
  \centering
  \begin{tabular}{lll}
    \toprule
         & $w=0.25$     & $w=0.35$  \\
    \midrule
    w/o similarity & 42.43  & 60.01    \\
    w/o decay    & 272.83 & 282.05     \\
    similarity $\&$ decay & \textbf{36.64} & \textbf{35.05}\\
    \bottomrule
  \end{tabular}
\end{table}

\begin{table}
  \caption{FID of the images that are applied with the constraints at the end of the interpolation.}
  \label{tbl:run_last}
  \centering
  \begin{tabular}{lll}
    \toprule
         & FID (Val.) & FID (Test)   \\
    \midrule
    FIT (SS) & 40.61 & 36.13     \\
    FIT (RA) & 264.64 & 197.25     \\
    \bottomrule
  \end{tabular}
\end{table}

\textbf{Potential applications to more scientific problems} -- Image interpolation has been used in other scientific domains such as biomedical imaging \cite{tavoosi2021} and satellite imagery in earth and space science \cite{vandal2021}. The concept of guided interpolation may be used to bring domain specific regularization into the interpolation generation process. The regularization could be visual pattern priors, structure or topology constraints, which would require modeling of appropriate regularizers subject to domain specific knowledge. While these are not included in this paper, we hope this work would motivate more use cases of guided interpolation in more science domains.

\textbf{Limitations}
Functionality and performance are not necessary reflected in visual patterns; for example, the material and the mass in internal facing parts. The proposed approach would not be able to tackle the types of regularization that cannot be well represented visually. In addition, the effectiveness of functional constraints is highly dependent on the appropriate surrogate models that can precisely synthesize the functional requirements and are computationally feasible.




\section{Conclusion}
We propose FIT to generate design interpolations subject to functional constraints. FIT generates wheel design drawings with high realism and with high conformance to functional constraint such as symmetry. We also found that strength of the functional constraint regularizer in generation has a tradeoff between realism and regularization, and the performance depends on the noise that a regularizer may produce. The framework may serve other types of functional constraints that can be  visually regularized. In the future we will explore more constraint regularizers beyond symmetry, for example, structure strength such as width of the spokes, the harmonic frequency, CFD requirements. These require that we have a surrogate model for each of those physical requirements.

\bibliographystyle{plain}
\bibliography{references}

\end{document}